\documentclass{article}
\usepackage{ijcai19}

\usepackage{times}
\usepackage{soul}
\usepackage{url}
\usepackage[hidelinks]{hyperref}
\usepackage[utf8]{inputenc}
\usepackage[small]{caption}
\usepackage{graphicx}
\usepackage{amsmath}
\usepackage{booktabs}
\usepackage{algorithm}
\usepackage{algorithmic}
\usepackage{bm}
\title{Weak Supervision Enhanced Generative Network for Question Generation}

\author{
Yutong Wang$^1$\and
Jiyuan Zheng$^1$\and
Qijiong Liu$^1$\and
Zhou Zhao$^1$\footnote{Zhou Zhao is the corresponding author.}\and
Jun Xiao$^1$\And
Yueting Zhuang$^1$
\affiliations
$^1$College of Computer Science and Technology, Zhejiang University, China\\
\emails
\{ytwang, jiyuanz, lqj, zhaozhou, yzhuang\}@zju.edu.cn,
junx@cs.zju.edu.cn
}

\begin{document}

\maketitle

\begin{abstract}
Automatic question generation according to an answer within the given passage is useful for many applications, such as question answering system, dialogue system, etc. Current neural-based methods mostly take two steps which extract several important sentences based on the candidate answer through manual rules or supervised neural networks and then use an encoder-decoder framework to generate questions about these sentences. These approaches neglect the semantic relations between the answer and the context of the whole passage which is sometimes necessary for answering the question. To address this problem, we propose the Weak Supervision Enhanced Generative Network (WeGen) which automatically discovers relevant features of the passage given the answer span in a weak supervised manner to improve the quality of generated questions. More specifically, we devise a discriminator, Relation Guider, to capture the relations between the whole passage and the associated answer and then the Multi-Interaction mechanism is deployed to transfer the knowledge dynamically for our question generation system. Experiments show the effectiveness of our method in both automatic evaluations and human evaluations.
\end{abstract}

\section{Introduction}
Question generation based on a passage (QG for short) is a challenging task that is described as generating a correct and fluent question for the answer according to a piece of text automatically. It requires thorough understanding of the context, including local entities, along with semantic features and relations in the text. Question generation task has drawn increasingly attention in natural language processing recently. Due to collections of large QA corpus~\cite{rajpurkar2016squad,nguyen2016ms,hermann2015teaching}, QG systems develop rapidly and apply successfully in many other tasks such as providing fruitful questions for question answering systems, generating exercises for students or automatically asking questions to get feedback in a conversation system~\cite{heilman2010good,du2017identifying}. However, labelling these datasets is expensive and usually requires mountains of manual labour. Besides, there are still strong biases  ~\cite{subramanian2017neural} in existing QA datasets including question types, linguistic styles and etc.

Asking questions is a universal technique to assess the reader's understanding of an article. Previous works tackled question generation using rule-based or template-based methods~\cite{heilman2010good,rus2010first} which analysed syntactic or linguistic features and transformed the text into questions. These approaches require human-designed features and rules that are not efficient and hard to transfer to a new domain. Besides, these traditional methods sometimes limit in fluent and human-like expression and cannot capture deeper semantic representation of the sentence.

Recent methods utilize the encoder-decoder neural networks to generate questions of the text and achieve state-of-the-art performances. Most of them use one or two sentences to generate questions. ~\cite{subramanian2017neural} firstly extract several potential candidate phrases using a sequence-to-sequence model and then feed these candidates into a encoder-decoder structure with an attention mechanism. ~\cite{du2017identifying} further encode paragraph-level information with an additional bidirectional LSTM to enhance the sub-task of key sentence selection and then use a global attention-based model to generate questions. These methods only extract several sentences for generating a question which discard the origin sequence of answer and necessary global information of the passage.

In this paper, we focus on end-to-end question generation of an answer given a corresponding passage. In order to leverage the information of the whole passage and to generate more answer-focused questions, we propose a novel end-to-end framework with weak supervision labels to enhance question generation. Different from the previous works, which learn to train a encoder-decoder model directly, we utilize weak supervision labels which are readily available to model the relations of a related passage and the answer span and then transfer the knowledge we learned to enhance question generation system. Specifically, a discriminator called Relation Guider is designed to distinguish whether the passage could provide necessary information to reach the answer or not. In other words, the Relation Guider aims at finding the valuable part of the passage which is helpful for comprehending the answer and identifying the deceptive passages that actually unnecessary for obtaining the answer. Subsequently, we design the Dual-channel Interaction module that contains multi-step modelling of the representation from the origin channel and channel of Relation Guider and a control gate to decide how much information is able to pass through from these channels respectively. Finally, we incorporate the Relation Guider and Dual-channel Interaction module into our encoder-decoder framework to predict questions. Note that our method is orthogonal to the lexical features enhancement like NER, POS~\cite{zhou2017neural}, which can be incorporated to further improve the performance.
We conduct detailed experiments to demonstrate the comparative performance of our approach. The contributions of this paper can be summarized as follows:

\begin{itemize}
    \item We design a weak supervision based discriminator called Relation Guider which models the answer and the corresponding passage jointly, to capture the relations between them and focus on the answer-related parts of the passage.
    \item We propose a Dual-channel Interaction mechanism that transfers the knowledge of the discriminator to guide the process of question generation.
    \item We propose the novel end-to-end Weak Supervision Enhanced Generative Network (WeGen) which utilizes the Relation Guider and combines the Dual-channel Interaction module into the encoder-decoder structure for question generation system.
    \item We conduct our experiments on the SQuAD dataset to examine and analyse the effectiveness of our framework on human evaluations and achieve considerable performance gain on all the automatic evaluation metrics.
\end{itemize}

\section{Related Work}
To economize manpower of generating questions, automatic question generation are utilized for question answering systems and dialogue systems. Some early studies are based on traditional methods that consider features of language for question generation. ~\cite{kunichika2004automated} convert declarative sentences into questions according to grammatical information. ~\cite{mitkov2003computer} propose a method that uses shallow rules and corpus of synonym to generate multiple-choice questions. ~\cite{khullar2018automatic} propose a syntax-based method which leverages dependency parsing to generate elegant questions. Pronouns and Adverbs are used in their framework and the system achieves high quality. ~\cite{flor2018semantic} parse the text in terms of POS, NER etc, and then process the text through the traditional pipelines to generate wh-questions and yes/no questions. ~\cite{rothe2017question} propose a statistic model that analyses syntax and probability of question types to learn how to produce creative and human-like questions. 

To improve the quality of question generation, ~\cite{subramanian2017neural} propose a two-step framework which extracts important phrases from the document and generates questions with an attention-based sequence-to-sequence model. ~\cite{du2017identifying} propose a similar framework which leverages neural model for selecting question-worthy sentences and generates questions with an encoder-decoder network. 

Some works benefit from end-to-end neural networks in computer vision and natural language processing.  ~\cite{zhang2017automatic} propose a neural image question generation model that learns image captions, image features and question types jointly. ~\cite{duan2017question} propose a retrieval-based model with convolution neural network and a recurrent neural network to obtain questions.  ~\cite{du2017learning} propose an sequence-to-sequence model that encodes sentence-level and paragraph-level information. ~\cite{sun2018answer} propose a neural model that considers embedding and positions of answers by concatenating answer position features to generate more relevant questions.


\begin{figure*}
  \centering
  \includegraphics[width=.85\textwidth]{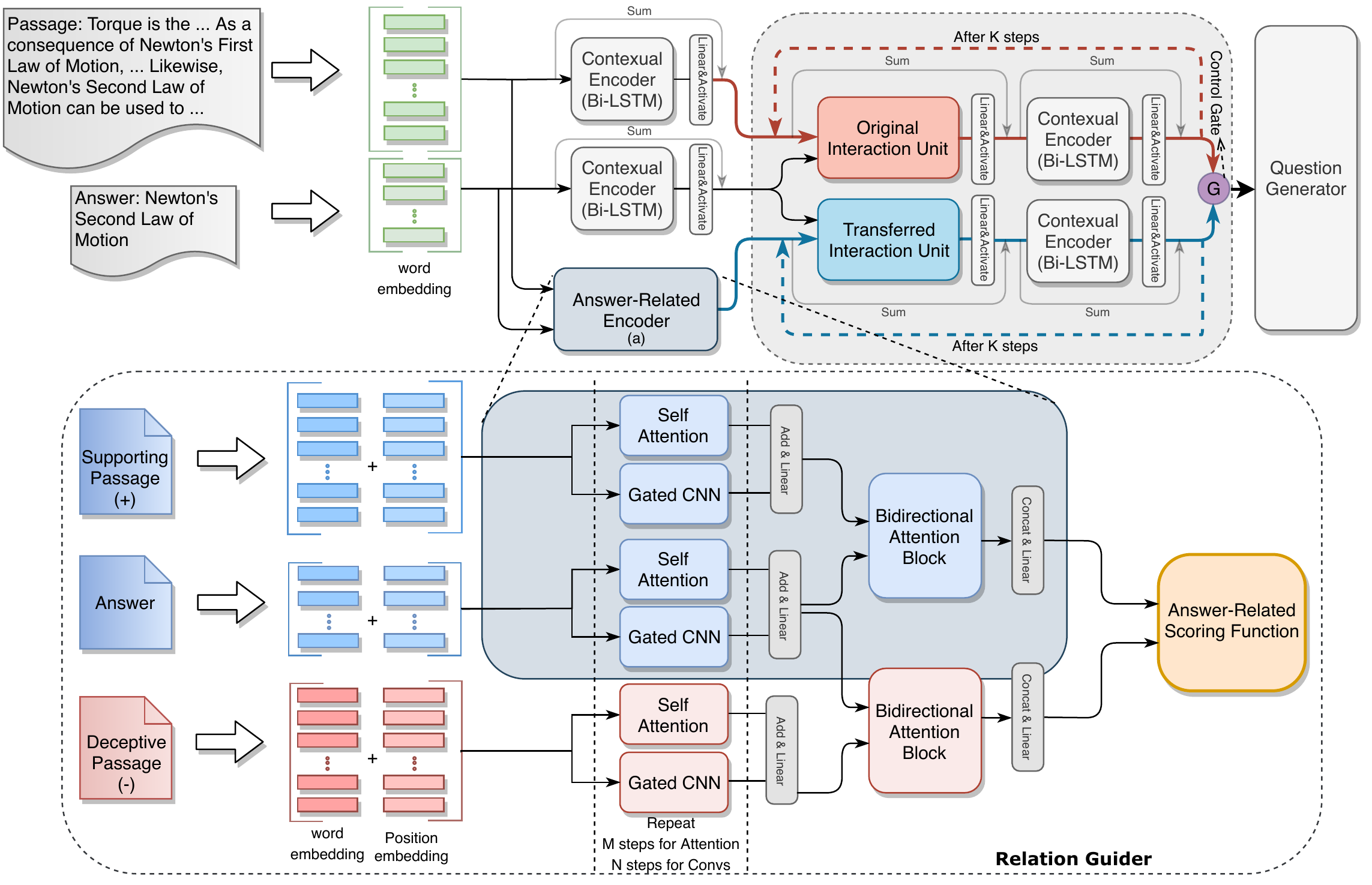}
  \caption{The entire framework of our Weak Supervision Enhanced Generative Network.}
  \label{qg-model}
\end{figure*}

\section{Our Approach}
In this section, we firstly define the question generation task and then propose our Weak Supervision Enhanced Generative Network(WeGen) for question generation. The architecture of our model is shown in Figure \ref{qg-model} which mainly contains four modules: (a)Relation Guider that pre-trains a novel weak supervised discriminator to learn relation between passages and answers, (b)Encoding Module that transfer the knowledge learned from Relation Guider to extract multi-channel features, (c)Multi-Interaction that fuses the multi-modal representations to capture the worthy content for the answer, (d)Decoding Module that generates questions with an attention-based decoder. 

\subsection{Problem Formulation}
In the question generation task, we are given a passage with n words $P={(p_1, p_2, ..., p_n)}$ and an answer $A={(a_1, a_2,..., a_m)}$ with $m$ words related to the passage. Our goal is to generate a question $\bar{Q}$ of an arbitrary length that satisfies:
\begin{equation}
    \bar{Q} = \arg\max_{Q} P(Q| P,A)
\end{equation}
where $P(Q| P,A)$ is the conditional log probability of the generated question given the passage and the answer. For a particular question, the equation is obtained as:
\begin{equation}
    log P(Q| P,A) = \sum_{t=0}^{K} log P(q_t | P,A,q_0,...,q_{t-1})
\end{equation}
where $K$ is the length of the generated question and $q_t$ is the $t$-th token of the question.

\subsection{Relation Guider}
\label{relation-guider}
In order to generating correct questions related to the answers, it is important to comprehend the passages, answers and the relation between them. We propose the Relation Guider to encode the passages and answers and learn the relation between them with only weak supervision labels. Labels are binary features which stand for the fact whether the answer could be obtained or inferred from the current passage or not.

In our Relation Guider, we firstly take the triplets $(answer, passage^+, passage^-)$ as input sequences, which are the answer, a supporting passage and a deceptive passage obtained from the search engine (See Details in Section \ref{ms-marco}). Then each word of the triplets is represented by pre-trained word embeddings. Due to the large amount of data and limits of computing resources, we propose the encoding method using convolutional neural networks and self-attention mechanism that support parallel computing to accelerate our pre-training process.
Considering the lack of the location information using convolutions and attentions, we follow ~\cite{vaswani2017attention} to use Position Embedding which encodes positions of tokens with $\sin$ and $\cos$ functions of different frequencies. The dimension of the Position Embedding is identical to word embedding and then we sum up these two embeddings.

Given the embeddings of the passage-answer triplets, Relation Guider encodes them respectively with gated convolutional neural networks and self-attention mechanism.
We stack multiple layers of the gated CNN~\cite{dauphin2017language} and self-attention units for better modelling the representations. The weights of the same layer are shared among the units of passages and answers while weight matrices of the different layers are not shared. Passages and answers are encoded as follows:
\begin{eqnarray}
    & \Tilde{\textbf{p}}  =  f([ SelfAtt(\textbf{p},\textbf{p},\textbf{p})]^M + [GatedCNN(\textbf{p})]^N) \\
    & \Tilde{\textbf{a}}  =  f([ SelfAtt(\textbf{a},\textbf{a},\textbf{a})]^M + [GatedCNN(\textbf{a})]^N) \\
\end{eqnarray}
where $[\cdot]^M$ and $[\cdot]^N$ stand for the repeat times of self-attention and convolution unit. Each $SelfAtt(\cdot)$ and $GatedCNN(\cdot)$ unit is wrapped with a residual mechanism, followed by a layer-normalization function. $f(\cdot)$ denotes a fully connected feed-forward layer with a $ReLU$ activation function.

Similar to the Bi-directional Attention Flow ~\cite{seo2016bidirectional}, we implement two-way attention mechanism to model the relations between the answer and current passage and pinpoint the important parts of the passage and attend to representations of the answer. Firstly, the similarity matrix $\bm{S} \in R^{n \times m}$ between the passage and answer is obtained by:
\begin{equation}
    \textbf{S}_{ij} = \bm{W_S}^{\top}[\Tilde{\textbf{p}}_i; \Tilde{\textbf{a}}_j; \Tilde{\textbf{p}}_i \odot \Tilde{\textbf{a}}_j]
\end{equation}
where $\bm{W_S}$ is a trainable matrix that computes the similarity between the $i$-th vector $\Tilde{\textbf{p}}_i$ of the passage and $j$-th vector $\Tilde{\textbf{a}}_j$ of the answer and $[;]$ is vector concatenation, $\odot$ is element-wise multiplication. 

Then the representations of the passage and answer interact in order to learn the passage-to-answer relation and answer-to-passage relation. We attend each token of passage to tokens of the answer and vice versa to get the attention vector between two tokens by:
\begin{eqnarray}
   & \hat{\textbf{p}}_i = \sum_{j} \textbf{S}_{ij} \ \cdot \  \Tilde{\textbf{a}}_j \\
   & \hat{\textbf{a}}_j = \sum_{i} \textbf{S}_{ij}\ \cdot \  \Tilde{\textbf{p}}_i
\end{eqnarray}
where the similarity matrix is normalized across the row and column when the weighted sum of $i$-th token and $j$-th token are calculated. 

To obtain the representations of two-way attention, the attended vectors of $P$ and $A$ are combined with the previous inputs by:
\begin{eqnarray}
   & \bar{\textbf{p}}_i = [\Tilde{\textbf{p}}_i;\hat{\textbf{p}}_i;\Tilde{\textbf{p}}_i \odot \hat{\textbf{a}}_j] \\
   & \bar{\textbf{a}}_j = [\Tilde{\textbf{a}}_j;\hat{\textbf{a}}_j;\Tilde{\textbf{a}}_j \odot \hat{\textbf{p}}_i]
\end{eqnarray}

After the interaction of $\Tilde{\textbf{p}}$ and $\Tilde{\textbf{a}}$, a feed-forward layer is employed to obtain the relevant scores of the current positive and negative passage-answer pairs. 
\begin{equation}
    s_{(p,a)} = Sigmoid(\bm{W^{(p,a)}}[\bar{\textbf{p}};\bar{\textbf{a}}])
\end{equation}
where $\bar{\textbf{p}},\bar{\textbf{a}}$ denote the outputs of the interaction layer. $\bm{W^{(p,a)}}$ is a trainable weight matrix and $s_{(p,a)}$ is the relevant score of the current passage and the answer. 

We then present the Answer-Related Scoring function learning method for judging relations between the answer and the passage. Given the output representations of passage-answers, we now design its loss function $\mathcal{L}_r$ as follows:
\begin{equation}
    \mathcal{L}_r = \sum_{i,j,k \in \mathcal{R}} max(0, c+s_{(p,a)}^{-}(a_i, p_j)-s_{(p,a)}^{+}(a_i, p_k))
\end{equation}
where the superscript $s_{(p,a)}^{+}(\cdot)$ and $s_{(p,a)}^{-}(\cdot)$ denote the  scores of "correct" passage-answer pair(that this answer could be answered by the passage) and "wrong" passage-answer pair (that we could not obtain this answer from the passage) for question answering. We define the hype-parameter $c (0 < c < 1 )$ controls the margin in the loss function and $\mathcal{R}$ is the set of passage-answer pairs.

\subsection{Encoder-Decoder Structure}
We then present the encoder of Weak Supervision Enhanced Generative Network for input encoding. Given the sequences of a passage-answer pair, we use the same pre-trained word embedding in Section \ref{relation-guider}. As shown in Figure \ref{qg-model}, the embeddings of $P$ and $A$ are encoded with two contextual encoder for which we choose bidirectional LSTM and the two encoders share weights in our question generation system. At the same time, the Answer-Related Encoder shown in Figure \ref{qg-model}(a) models the passage and answer and outputs the answer-related passage features. Then these three contextual representations(origin passage, answer, answer-related passage) flow in Dual-channel Interactions Module to learn variant features of relations and contextual representations.
This Module includes two channels: the original interaction path and our transferred interaction path. The operations of two paths are mirrored including an interaction unit modelling the passage-answer pair and an Bi-LSTM layer further encoding the current representations. It is worth noting that the transferred interaction unit takes the semantic features of passage pre-trained on our Relation Guider as inputs, which is already able to learn the relations between $P$ and $A$ and capture associated information of the passage. 

The operations of original interaction unit and transferred interaction unit are identical to the bidirectional attention block, followed by a linear layer and ReLU activation function. In the meantime, each unit including the contextual encoder is wrapped with residual mechanism. And then we bind representations of origin passage and the answer, representations of answer-aware passage and answer through the pipelines of variant interactions for repeating K steps. The weights of Answer-Related Encoder are fixed during our training process. 

Then we propose a control gate to integrate the knowledge learned in the original encoder and Relation Guider dynamically that is defined as:
\begin{eqnarray}
   & \bm{g} = \sigma(\bm{W_g}[\bm{x};\bm{y}]+\bm{b_g}) \\
   & \bm{G} = \bm{g} \cdot \bm{x} + (\bm{1}-\bm{g}) \cdot \bm{y}
\end{eqnarray}
where $\bm{W_g}$ and $\bm{b_g}$ are the trainable parameters and $x,y$ symbolize the representations of passage from two different channels. We define $\bm{g}$ as the control gate and $\bm{G}$ is the joint representations of the passage. $\cdot$ is element-wise matrix multiplication and $[;]$ is operation of concatenation.

During the decoding procedure, we use attention-based LSTM decoder with copy mechanism to deal with the out-of-vocabulary problem.
The encoder attention memory is the joint representations of the passage $\bm{G}$. 
Then at each decoding time step, the attentive results and the previous word embedding are injected into our decoder unit. The attentive outputs of the encoder are used for initializing the decoder LSTM hidden state.

\section{Experiments}
\subsection{Datasets}
\label{ms-marco}
The \textbf{MS MARCO} dataset is a large scale dataset collected from Bing. This dataset contains 1,010,916 questions and 8,841,823 related passages extracted from 3,563,535 web documents. Due to a large quantity of question-answer pairs, we are able to use MS MARCO to train our discriminator. We notice that each question-answer pair is followed with 10 passages obtained from the search engine and some passages are not necessary to answer the current question. We denote these passages as negative samples and the passages that provide supporting facts for answering the question are positive samples. Thanks to these implicit label information of the MS MARCO dataset, we could construct our samples for pre-training the Relation Guider as follows: 
\begin{equation}
    \bm{S} = \textit{Set}(a,p^{+},p^{-})
\end{equation}
The \textbf{SQuAD} dataset ~\cite{rajpurkar2016squad} is one of the most influential reading comprehension datasets which contains over 100k questions of 536 Wikipedia articles created by crowd-workers and the answers are continuous spans in the passages.
Following ~\cite{du2017learning}, we conduct our experiments on the train and development sets which are visible for us. The whole dataset(train set and dev set) is randomly divided into a training set (80\%), a development set (10\%) and a test set (10\%) at the article level.

\subsection{Implementation Details}
We extract vocabulary from the training set and keep the $45000$ most frequent words for source sequence and target sequence, and other tokens are replaced by a special \textit{@UNK@} token. 
The word embeddings are initialized from GloVe~\cite{pennington2014glove} pretrained  embeddings and are fixed during the training time.

For model hyperparameters, we have 4 convolutional filters with kernel size 7 for all convolutional blocks and all the self attention blocks are Multi-head Attention~\cite{vaswani2017attention} with 8 attention heads.
We adopt Adam optimizer with a learning rate of $0.001$, $\beta_1=0.9$ and $\beta_2=0.999$ and batch size is set to 16.
we train the model for a maximum of 20 epochs and use early stopping with the patience set to 5 epochs according to the BLEU-4 score on validation set.
At the validation and test time, we use beam search with beam size set to 5.

\subsection{Baselines}
\begin{itemize}
    \item \textbf{Vanilla Seq2Seq}. ~\cite{sutskever2014sequence} propose a basic sequence-to-sequence model which encodes the input sequence of variable length with recurrent neural networks and decodes the target sequence. The embeddings of answers are concatenated into the passages representations.
    \item \textbf{Attention-Seq2Seq}. Attention-based Seq2Seq neural network extends the Vanilla Seq2Seq with attention mechanism. The model reads the attentive outputs of the encoder at every decoding step and transport the results to the decoder unit.
    \item \textbf{Seq2Seq+Attention+Copy Mechanism}. We carefully implemented the neural sequence-to-sequence model with attention and copying mechanism.
    \item \textbf{Transformer}. Transformer~\cite{vaswani2017attention} is a totally attention-based sequence-to-sequence model without using RNN cells. The basic block includes a Multi-Head Attention layer and a fully connected feed-forward layer. A Multi-Head Attention layer contains several attention units and the outputs of all units are concatenated and mapped to the size of input.
\end{itemize}

\subsection{Automatic Evaluation}
\label{auto-eval}
Table \ref{tab:metrics} shows different metric scores of baselines and our model. The experimental results reveal a number of interesting points. The copy mechanism improve the results significantly. It uses attentive read from word embedding of sequence in encoder and selective read from location-aware hidden states to enhance the capability of decoder and proves effectiveness of the repeat pattern in human communication.
Transformer structure performs badly and just achieves better results than Vanilla Seq2Seq. This suggests that the pure attention-based models are not sufficient for question generation and the local features of sequences and the variant semantic relations should be modelled more effectively.

In all cases, our WeGen achieves the best performance, which shows that leveraging both our novel pre-trained relation extractor via weak supervision labels and multiple interactive generator can improve the performance of QG system significantly. There are two novel parts of our WeGen that play a vital role in our question generation. The Relation Guider learns semantic matching patterns between the passages and the answers, and our Dual-channel Interaction mechanism captures the information transferred from the Relation Guider and models the variant interactions between dual-channel representations of passages and answers, in order to further enhance our encoding process. And the ablation results of WeGen are obviously better than WeGen without transferring knowledge in our QG system which further proves the importance of our transfer learning module.

\begin{table*}[ht]
\centering
\begin{tabular}{l|cccc|c|c}  
\toprule
Model  & BLEU 1 & BLEU 2 & BLEU 3 & BLEU 4 & ROUGE-L & METEOR\\
\hline
\hline
Vanilla Seq2Seq  & 17.13  & 8.28 & 4.74 & 2.94  & 18.92 & 7.23   \\
Seq2Seq+Attention  & 17.90  & 9.64 & 5.68 & 3.34  & 19.95 & 8.63   \\
Transformer  & 15.14  & 7.27 & 3.94 & 1.61  & 16.47 & 5.93  \\
Seq2Seq+Attention+Copy & 29.17  & 19.45 & 12.63 & 10.43  & 28.97 & 17.63\\ \hline
\textbf{WeGen}  & \textbf{32.65}  & \textbf{22.14} & \textbf{15.86} & \textbf{12.03}  & \textbf{32.36} & \textbf{20.25} \\
WeGen w/o pre-training & 31.14 & 20.26 & 13.87 & 11.25& 31.04 & 18.42 \\ 
\bottomrule
\end{tabular}
\caption{Comparison with other methods on SQuAD dataset. We demonstrate automatic evaluation results on BLEU 1-4, ROUGE-L, METEOR metrics. The best performance for each column is highlighted in boldface. The WeGen without pre-training means the pipeline of Answer-Related Encoder and Transferred Interaction module are not used and the control gate is abandoned. Refer to Section \ref{auto-eval} for more details.}
\label{tab:metrics}
\end{table*}

\subsection{Human Evaluation}
\begin{table}
\centering
\begin{tabular}{p{0.9\columnwidth}}
\toprule
\textbf{Passage}: Civil disobedience is usually defined as pertaining to a citizen's relation to the state and its laws, as distinguished from a constitutional impasse in which two public agencies... \\
\textbf{Answer}: constitutional impasse \\
\textbf{Reference Question}: What does not constitute as civil disobedience? \\
\textbf{Question from Seq2Seq+Attention}: What is an example of a party in the US? \\
\textbf{Question from Seq2Seq+Attention+Copy}: What is the term used to describe Civil disobedience? \\
\textbf{Question from WeGen}: Civil disobedience is distinguished from what? \\
\bottomrule
\end{tabular}
\caption{A case from test set shows the comparison of questions generated from baselines and our model WeGen.}
\label{tab:case}
\end{table}

\begin{figure}
\begin{minipage}[t]{0.5\linewidth}
\centering
\includegraphics[width=\textwidth]{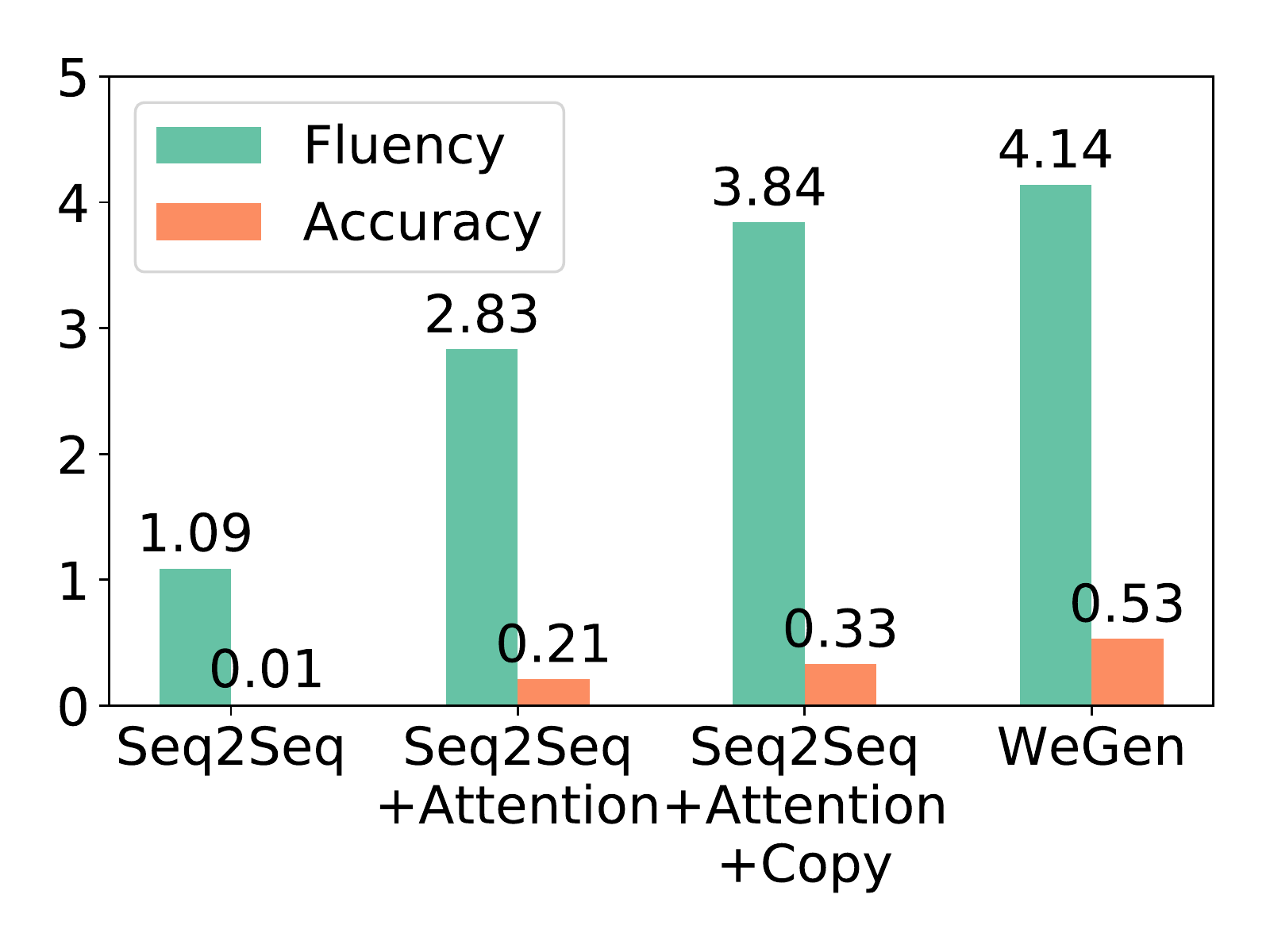}
\caption{Fluency and Accuracy results.}
\label{fig:side:a}
\end{minipage}%
\begin{minipage}[t]{0.5\linewidth}
\centering
\includegraphics[width=\textwidth]{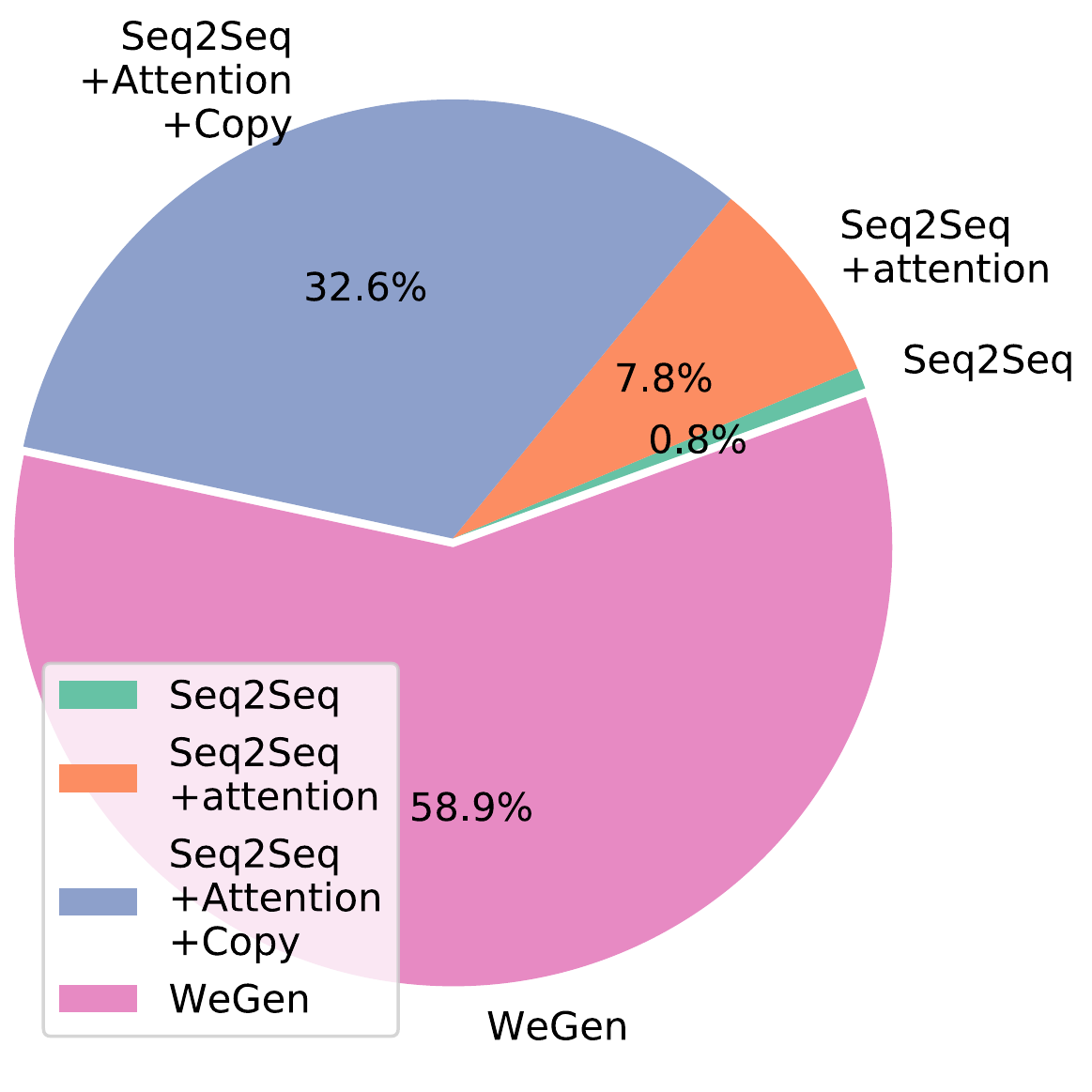}
\caption{Turing@1 Results.}
\label{fig:side:b}
\end{minipage}
\end{figure}

We introduce three human evaluation metrics to verify the performance of baselines and our model.

\begin{itemize}
\item \textbf{Fluency}. We follow the evaluation method in machine translation to evaluate the language fluency of generated questions. The scale is from 1 to 5 denoted as follows:(1) Incomprehensible target question, (2) Incoherent target question,(3) Non-native kind of target question, (4) Good quality target question, (5) Flawless target question.
\item \textbf{Accuracy}. Accuracy is designed to verify whether the answer can be replied to the generated question or not. It is calculated from the proportion of the correctly answered questions.
\item\textbf{Turing@1}. We propose the Turing@1 metric which ranks all the questions generated by different models without considering the answer from the perspective of human-like expression as a Turing Test. It computes the frequency that the current question from a specific model ranks top among all the questions over the participants via:
\begin{equation}
    Turing@1 = \frac{\sum_{N} f(x)}{N}, f(x)=
\begin{cases}
1& \text{x=Ranking Top}\\
0& \text{Others.}
\end{cases} 
\end{equation}
where $N$ is the number of participants.
\end{itemize}

Figure \ref{fig:side:a} and \ref{fig:side:b} shows the scores of three human evaluation methods. We conduct our human evaluations on 30 participants over 100 samples from test set. Though the amount of samples limits, we could still peek some learning patterns of different question generation methods. WeGen obtains the best language quality which should thanks to multi-stage contextual modelling of text. And the results of Accuracy achieves $0.53$ of our WeGen much better than other baselines, which reveals that our WeGen could generate more answer-related questions due to our answer-passage Relation Guider. Figure \ref{fig:side:b} shows the proportion of the participants that judge the question generated by a specific model closest to human query characteristics. The WeGen obtains most Top1 times among all the models and we find the reason that other baselines such as Seq2Seq+Attention or Seq2Seq+Attention+Copy will repeat meaningless words which are easily distinguished by human.

\subsection{Ablation and Case Study}

\begin{figure}[t]
  \centering
  \includegraphics[width=.42\textwidth]{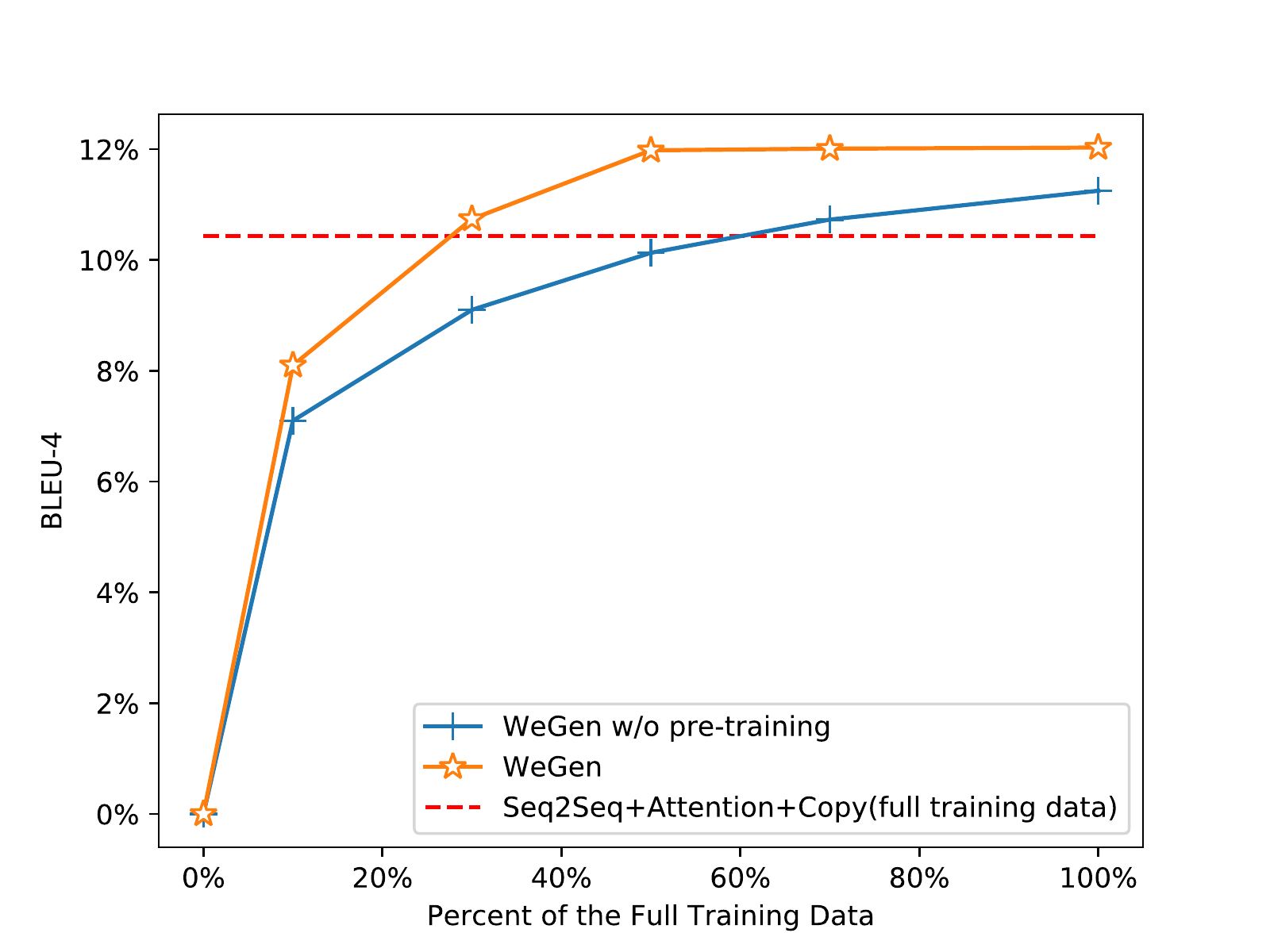}
  \caption{The performance comparison of our model on variant proportions of training data.}
  \label{ablation-study}
\end{figure}

To study the effectiveness of pre-training, we extract \{$10\%, 30\%, 50\%, 70\%$\} of the full training data and train WeGen and WeGen-without-pre-training on these different portions of training data respectively.
The results are demonstrated in Figure \ref{ablation-study}. 

We can find that WeGen trained on just 30\% of original training data surpasses the BLEU-4 score of Seq2Seq+Attention+Copy trained on full training data(the red dotted line). Meanwhile, WeGen also outperforms WeGen-without-pre-training by 1-2 percent on 30\%, 50\%, 70\% of training data. And it is reasonable that the performance gained by pre-training are limited when training data increases because the modeling capacity also increases.  

Table \ref{tab:case} shows an example in our experiments. In contrast to our model, Seq2Seq model proposes a question unrelated with the passage either the answer, while Seq2Seq+Attention+Copy model just proposes a question related to passage but incoherent with the answer. However, our model skips the long attributive adjunct and captures the correct relationships with verbs and subjects to propose the reasonable question about the answer, which is similar with the reference in semantics.

\section{Conclusion}
In this paper we have proposed a novel method for question generation called Weak Supervision Enhanced Generative Network. This approach firstly leverages the easily reachable labels to train a discriminator for matching passage-answer pairs, in order to capture the semantic relations between a passage and an related answer. And then we design our question generator with the Multi-Interaction method to transfer the knowledge of this discriminator dynamically, in consideration of obtaining more fine-grained information, such as the answer-related part of the passage and other potential relations. The experimental results show that the effectiveness of our approach in the automatic evaluations and human evaluations. 

\section*{Acknowledgments}
This work was supported by the National Natural Science Foundation of China under Grant No.61602405, No.U1611461, No.61751209, No.61836002. This work was also supported by the China Knowledge Centre of Engineering Sciences and Technology, Joint Research Program of ZJU and Hikvision Research Institute and the Alibaba Innovative Research.
\appendix

\clearpage
\bibliographystyle{named}
\bibliography{ijcai19}

\end{document}